\documentclass[wcp]{jmlr}

\usepackage{graphicx}
\usepackage{microtype}
\usepackage{graphicx}
\usepackage{float}
\usepackage{mathtools}
\usepackage{enumitem}
\usepackage{booktabs} 
\usepackage{amsmath}
\usepackage{amssymb}
\usepackage{fixltx2e}
\usepackage{enumitem}
\usepackage{algorithm}
\usepackage[noend]{algpseudocode}
\setlist{nolistsep}

\usepackage{mathtools} 
\usepackage{booktabs} 
\usepackage{tikz} 

\def\NoNumber#1{{\def\alglinenumber##1{}\State #1}\addtocounter{ALG@line}{-1}}

\usepackage{enumitem}
\usepackage{hyperref}
\usepackage{afterpage}

\usepackage[american]{babel} 

\makeatletter
\let\Ginclude@graphics\@org@Ginclude@graphics 
\makeatother

\jmlrvolume{157}
\jmlryear{2021}
\jmlrworkshop{ACML 2021}

\title[Unfair Edge Prioritization and Discrimination Removal]{A Causal Approach for Unfair Edge Prioritization and Discrimination Removal}

  \author{\Name{Pavan Ravishankar} \Email{pavan.rshankar@gmail.com}\\
   \Name{Pranshu Malviya} \Email{cs19s031@cse.iitm.ac.in}\\
   \Name{Balaraman Ravindran} \Email{ravi@cse.iitm.ac.in}\\
   \addr Robert Bosch Center for Data Science and Artificial Intelligence, IIT Madras, Chennai, India\\
   \addr Department of Computer Science and Engineering, IIT Madras, Chennai, India}

\editors{Vineeth N Balasubramanian and Ivor Tsang}

\begin{document}
\allowdisplaybreaks
\maketitle

\begin{abstract}
In budget-constrained settings aimed at mitigating unfairness, like law enforcement, it is essential to prioritize the sources of unfairness before taking measures to mitigate them in the real world. Unlike previous works, which only serve as a caution against possible discrimination and de-bias data after data generation, this work provides a toolkit to mitigate unfairness during data generation, given by the \textit{Unfair Edge Prioritization} algorithm, in addition to de-biasing data after generation, given by the \textit{Discrimination Removal} algorithm. We assume that a non-parametric Markovian causal model representative of the data generation procedure is given. The edges emanating from the sensitive nodes in the causal graph, such as race, are assumed to be the sources of unfairness. We first quantify \textit{Edge Flow} in any edge $X \rightarrow Y$, which is the belief of observing a specific value of $Y$ due to the influence of a specific value of $X$ along $X \rightarrow Y$. We then quantify \textit{Edge Unfairness} by formulating a non-parametric model in terms of edge flows. We then \textit{prove} that cumulative unfairness towards sensitive groups in a decision, like race in a bail decision, is non-existent when edge unfairness is absent. We prove this result for the non-trivial non-parametric model setting when the cumulative unfairness cannot be expressed in terms of edge unfairness. We then measure the \textit{Potential to mitigate the Cumulative Unfairness} when edge unfairness is decreased. Based on these measurements, we propose the \textit{Unfair Edge Prioritization} algorithm that can then be used by policymakers. We also propose the \textit{Discrimination Removal Procedure} that de-biases a data distribution by eliminating optimization constraints that grow exponentially in the number of sensitive attributes and values taken by them. Extensive experiments validate the theorem and specifications used for quantifying the above measures.\\

\end{abstract}

\section{INTRODUCTION}
\textbf{Motivation and Problem:} Anti-discrimination laws in the U.S. prohibit unfair treatment of people based on sensitive features, such as gender or race \citep{act1964civil}. The fairness of a decision process is based on \textit{disparate treatment} and \textit{disparate impact}. Disparate treatment, referred to as intentional discrimination, is when sensitive information is explicitly used to make decisions. Disparate impact, referred to as unintentional discrimination, is when decisions hurt certain sensitive groups even when the policies are neutral. For instance, only candidates with a height of $6$ feet and above are selected for basketball teams. This might eliminate players of a certain race. Unjustifiable disparate impact is unlawful \citep{barocas2016big}. In high-risk decisions, such as in the criminal justice system, it is imperative to mitigate unfairness resulting from either disparate treatment or disparate impact. Considering that agencies operate in a budget-constrained scenario owing to limited resources, it is essential to prioritize potential sources of unfairness before we take measures to mitigate them. This paper proposes the Unfair Edge Prioritization methodology for prioritizing these sources before mitigating unfairness during the data generation phase. Further, this paper also proposes the Discrimination Removal procedure to de-bias data distribution after data is generated.  \\

\begin{figure}[t!]
\centering
\includegraphics[width=0.24\linewidth, height=3cm] {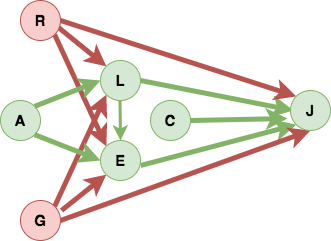}
\caption{Bail Decision Causal Graph. Each node is generated from its parents. \textbf{Edges:} Fair Edges in Green, Unfair Edges in Red; \textbf{Nodes:} Sensitive nodes in Red, Not sensitive nodes in Green; R: Race; A: Age, G: Gender, L: Literacy, E: Employment, C: Case characteristics, J: Judicial Bail decision} \label{main_graph}
\end{figure}

We motivate our problem through the following illustration using Fig. \ref{main_graph}. Consider the problem of reducing unfairness in the \textit{bail decision} $J$ towards a specific \textit{racial} group $R$. We use an unfair edge as the potential source of unfairness as in \cite{chiappa2018causal}. An unfair path contains at least one unfair edge. The unfairness propagates along all the unfair paths from the racial group $R$ to the bail decision $J$. Although discrimination has been quantified in previous works \citep{zhang2017causal}, it only serves as a caution against possible discrimination. There is utility when such notes of caution, like ``discrimination exists in the bail decision $J$ towards the racial group $R$", are augmented with tangible information to mitigate discrimination, like ``unfairness in the unfair edge $R \rightarrow L$ is responsible for discrimination in the bail decision $J$ towards the racial group $R$". Then, the agencies can attempt to address the real-world issues underlying $R \rightarrow L$, such as lack of scholarships for racial group $R$. The challenge lies in providing such tangible information and methodologies for mitigating discrimination such as the amount of unfairness present in an edge, the measure of how edge unfairness affects cumulative unfairness \footnote{Cumulative unfairness captures discrimination due to unequal influences of $R=r$ on $J=j$ as compared to $R=r'$ on $J=j$ via the directed paths from $R$ to $J$.}, prioritizing the unfair edges, and removing discrimination. \\ 

This paper attempts to provide such ``tangible" information using Pearl's framework of Causal Inference \citep{pearl2009causality}. The contributions of this paper are as follows,\\
\begin{enumerate}[nosep]
    \item \textit{\textbf{Quantify Edge Flow}} in any edge $X \rightarrow Y$, which is the belief of observing a specific value of $Y$ due to the influence of a specific value of $X$ along the edge $X \rightarrow Y$.
    \item \textit{\textbf{Quantify Edge Unfairness}} in any edge $X \rightarrow Y$, which is the average difference in conditional probability of $Y$ given its parents $Pa(Y)$, $\mathbb{P}(Y|Pa(Y))$, with and without edge flow along $X \rightarrow Y$. It measures average unit contribution of edge flow in $X \rightarrow Y$ to $\mathbb{P}(Y|Pa(Y))$. We formulate a non-parametric model for $\mathbb{P}(Y|Pa(Y))$ in terms of edge flows along the parental edges of $Y$ (see Theorem \ref{cptdecom}).
    \item \textit{\textbf{Prove}} that the discrimination in any decision towards any sensitive groups is non-existent when edge unfairness is eliminated. The proof is non-trivial in the non-parametric model setting of \textit{CPT}s as cumulative unfairness cannot be expressed in terms of edge unfairness. We derive this result by upper bounding the absolute value of cumulative unfairness and showing that the upper bound becomes zero when edge unfairness is zero (see Theorem \ref{boundtheorem} and Corollary \ref{boundtheoremcorol}).
    \item \textit{\textbf{Quantify the Potential to Mitigate Cumulative Unfairness}} by calculating the derivative of the upper bound w.r.t edge unfairness. We do this as cumulative unfairness cannot be expressed in terms of edge unfairness in a non-parametric model setting of \textit{CPTs}.
    \item \textit{\textbf{Propose an Unfair Edge Prioritization}} algorithm to prioritize unfair edges based on their potential to mitigate cumulative unfairness and edge unfairness. Using these priorities, agencies can address the real-world issues underlying the unfair edge with the top priority.
    \item \textit{\textbf{Propose a Discrimination Removal algorithm}} to de-bias data distribution by eliminating exponentially growing constraints and subjectively chosen threshold of discrimination. \\
\end{enumerate}

\noindent \textbf{Contents:} We discuss the preliminaries in Section \ref{preliminaries}; quantify edge flow, edge unfairness, its impact on cumulative unfairness, and prove that discrimination is absent when edge unfairness is eliminated in Section \ref{section4}; propose unfair edge prioritization and discrimination removal algorithm in Section \ref{algorithms}; discuss experiments in Section 6; discuss related work in Section \ref{section4}; discuss conclusion in Section \ref{conclusion}.
\section{PRELIMINARIES}
\label{preliminaries}
Throughout the paper, we use boldfaced capital letters $\mathbf{X}$ to denote a set of nodes; italicized capital letter $X$ to denote a single node; boldfaced small letters $\mathbf{x}$ to denote a specific value taken by $\mathbf{X}$; italicized small letter $x$ to denote a specific value taken by its corresponding node $X$. $\mathbf{x}_{\mathbf{A}}$ restricts the values of $\mathbf{x}$ to the node-set $\mathbf{A}$. $Pa(X)$ denotes parents of $X$ and $pa(X)$ the specific values taken by them. Each node is associated with the conditional probability table $\mathbb{P}(X|Pa(X))$.\\

\noindent \textbf{Assumptions:} We assume the following are given: (1) A \textit{Markovian causal model} $M$ consisting of a causal graph $\mathbb{G}= (\mathbf{V,E})$ with variables $V$ and edges $E$, (2) Conditional probability tables (\textit{CPT}s) $\mathbb{P}(V|Pa(V))$ where $\forall V \in \mathbf{V}$ and $Pa(V)$ are the parents of $V$, (3) A list of \textit{sensitive nodes}, whose emanating edges are potential sources of unfairness, (4) The variables in the causal graph are \textit{discrete and observed}. Even though the theorem results extend to the continuous variable setting, Assumption (4) is made to not digress into inference and identifiability challenges \citep{avin2005identifiability}. This paper does not make any assumptions about the deterministic functions in the causal model.

\begin{definition}
\textbf{Node interventional distribution} denoted by $\mathbb{P}(\mathbf{Y}|do(\mathbf{X}=\mathbf{x}))$ is the distribution of $\mathbf{Y}$ after forcibly setting $\mathbf{X}$ to $\mathbf{x}$ irrespective of the values taken by the parents of $X$ \citep{pearl2009causality}.
\end{definition}
\begin{definition}
A \textbf{causal model} is formally defined as a triple $M=<\mathbf{U},\mathbf{V},\mathbb{F}>$ where,
\begin{enumerate}
    \item $\mathbf{U}$ is a set of unobserved random variables also known as exogenous variables that are determined by factors outside the model. A joint probability distribution $\mathbb{P}(\mathbf{u})$ is defined over the variables in $\mathbf{U}$.
    \item $\mathbf{V}$ is a set of observed random variables also known as endogenous that are determined by variables in the model, namely, variables in $\mathbf{U} \cup \mathbf{V}$. 
    \item $\mathbb{F}$ is a set of deterministic functions $\{f_1,... ,f_i, ...\}$ where each $f_i$ is a mapping from $\mathbf{U} \times (\mathbf{V} \backslash X_i)$ to $X_i$ written as,
    \begin{align}
        x_i= f_i(pa(X_i), \mathbf{u}_i)
    \end{align}
    where $X_i \in \mathbf{V}$, $pa(X_i)$ are the specific values taken by the observed set of parents of $X_i$ and $\mathbf{u}_i$ are the specific values taken by unobserved set of parents of $X_i$. \\
\end{enumerate}
Each causal model is associated with a causal graph $\mathbb{G}=(\mathbf{V},\mathbf{E})$ where $\mathbf{V}$ are the observed nodes and $\mathbf{E}$ are the directed edges. We assume that the causal model is Markovian which means that all exogenous variables $\mathbf{U}$ are mutually independent and each node is independent of its non-descendants conditional on all its parents. For a markovian model joint distribution $\mathbb{P}(\mathbf{V})$ is given by,
\begin{align}
    \mathbb{P}(\mathbf{V}) = \underset{V\in\mathbf{V}}{\prod} \mathbb{P}(v|pa(V))
\end{align}
where $\mathbb{P}(v|pa(V))$ is the conditional probability table \textit{CPT} associated with $V$.
\end{definition}


\begin{definition}
\textbf{Identifiability:}
\label{section:iden}
Let $\mathbb{G}$ be a causal graph. A node interventional distribution $\mathbb{P}(\mathbf{Y}|do(\mathbf{X}=\mathbf{x}))$, i.e., probability of $\mathbf{Y}$ when $\mathbf{X}$ is forcibly set to $\mathbf{x}$ is said to be \textit{identifiable} if it can be expressed using the observational probability $\mathbb{P}(\mathbf{V})$. When $\mathbb{G}$ comprises of only observed variables as in our work,
    \begin{align}
    \mathbb{P}(\mathbf{y}|do(\mathbf{x})) &=\underset{\mathbf{v}\in\mathbf{V\backslash \{X,Y\}}} {\sum}~~\underset{V \in \mathbf{V} \backslash\{\mathbf{X,Y}\},\mathbf{Y=y}}{~\prod}\mathbb{P}(v|pa(V))|_{\mathbf{x}} \label{factorization}
    \end{align}
\end{definition}
\begin{definition}
\textbf{Total Causal Effect}
\label{te}
$TE_{\mathbf{y}}(\mathbf{x}_2, \mathbf{x}_1)$ measures causal effect of variables $\mathbf{X}$ on decision variables $\mathbf{Y=y}$ when it is changed from $\mathbf{x}_1$ to $\mathbf{x}_2$ written as, 
\begin{align}
TE_{\mathbf{y}}(\mathbf{x}_2, \mathbf{x}_1) = \mathbb{P}(\mathbf{y}|do(\mathbf{x}_2)) - \mathbb{P}(\mathbf{y}|do(\mathbf{x}_1))
\end{align}
\end{definition}
\begin{definition}
\label{pse}
\textbf{Path-specific effect} $SE_{\pi, \mathbf{y}}(\mathbf{x}_2,\mathbf{x}_1)$ measures effect of node $\mathbf{X}$ on decision $\mathbf{Y=y}$ when it is changed from $\mathbf{x}_1$ to $\mathbf{x}_2$ along the directed paths $\boldsymbol{\pi}$, while retaining $\mathbf{x}_1$ 
for the directed paths not in $\boldsymbol{\pi}$ i.e. $\tilde{\boldsymbol{\pi}}$ written as,
\begin{align}
SE_{\pi,\mathbf{Y=y}}(\mathbf{x}_2, \mathbf{x}_1) = \mathbb{P}(\mathbf{y}|do(\mathbf{x}_2|_{\pi}, \mathbf{x}_1|_{\tilde{\boldsymbol{\pi}}}))- \mathbb{P}(\mathbf{y}|do(\mathbf{x}_1))
\end{align}
\end{definition}

\begin{definition}
A trail $V_1 \rightleftharpoons .....\rightleftharpoons V_n$ is said to be an \textbf{active trail} given a set of nodes $\mathbf{X}$ in $\mathbb{G}$ if for every v-structure $V_i\rightarrow V_j \leftarrow V_k$ along the trail, $V_j$ or any descendent of $V_j$ is in $\mathbf{X}$ and no other node in the trail belongs to $\mathbf{X}$. \label{activetrail}
\end{definition}
 
\begin{definition}
$\mathbf{A}$ is said to be \textbf{d-separated} from $\mathbf{B}$ given $\mathbf{C}$ in a graph $\mathbb{G}$ $(d\text{-}sep_{\mathbb{G}} (\mathbf{A};\mathbf{B}|\mathbf{C}))$ if there is no active trail from any $A \in \mathbf{A}$ to any $B \in \mathbf{B}$ given $\mathbf{C}$ as discussed in \cite{pearl2009causality}, and \cite{koller2009probabilistic} . If there is atleast one active trail from any $A \in \mathbf{A}$ to any $B \in \mathbf{B}$ given $\mathbf{C}$, then $\mathbf{A}$ is said to be \textbf{d-connected} from $\mathbf{B}$ given $\mathbf{C}$ in a graph $\mathbb{G}$ $(d\text{-}conn_{\mathbb{G}} (\mathbf{A}; \mathbf{B}|\mathbf{C}))$ as shown in Fig. 1.3 in \cite{pearl2009causality}.
\end{definition}

\begin{theorem}
If sets $\mathbf{X}$ and $\mathbf{Y}$ are d-separated by $\mathbf{Z}$ in a DAG $\mathbb{G}(\mathbf{E,V})$, then $\mathbf{X}$ is independent of $\mathbf{Y}$ conditional on $\mathbf{Z}$ in every distribution $\mathbb{P}$ that factorizes over $\mathbb{G}$. Conversely, if $\mathbf{X}$ and $\mathbf{Y}$ are d-connected by $\mathbf{Z}$ in a DAG $\mathbb{G}$, then $\mathbf{X}$ and $\mathbf{Y}$ are dependent conditional on $\mathbf{Z}$ in at least one distribution $\mathbb{P}$ that factorizes over $\mathbb{G}$ as shown in Theorem 1.2.4 in \cite{pearl2009causality}. \label{dseptheorem}
\end{theorem}

\noindent \textbf{Unfair edge $S \rightarrow X$:} Unfair edge is a directed edge $S \rightarrow X$ with $S$ being a sensitive node like race. Set of unfair edges in $\mathbb{G}$ is denoted by $\mathbf{E}^{\text{unfair}}_{\mathbb{G}}$. Unfair edge is a potential source of unfairness. For instance, in Fig. \ref{main_graph}, $G\,\to\,E$ is unfair if the accused is denied admission to co-ed institutions based on gender. On the other hand, $G\,\to\,E$ is fair, if only gender-specific institutions existed in the locality as discussed in \cite{chiappa2018causal}. Hence, the usage of the term \textit{potential}.\\

\noindent \textbf{Unfair paths $\boldsymbol{\pi}^{\text{unfair}}_{\mathbf{S},Y,\mathbb{G}}$:} Unfair paths are the set of directed paths from sensitive node $S \in \mathbf{S}$ to the decision node $Y$ in graph $\mathbb{G}$. Unfair paths capture how unfairness propagates from the sensitive nodes onto a destination node. For instance, in Fig. \ref{main_graph}, $\boldsymbol{\pi}^{\text{unfair}}_{G,J,\mathbb{G}}$ consists of $G\,\to\,E\,\to\,J$ that captures how unfairness in the edge $G\,\to\,E$ propagates to $J$. Non-causal paths do not propagate unfairness from sensitive nodes. Suppose there is another node, say religious belief $R$, and another non-causal path, say $R\,\leftarrow\,E\,\to\,J$. Still, $R\,\leftarrow\, E\,\to\,J$ is fair because bail decision $J$ is taken based on employment $E$ and not on religious belief $R$ as discussed in \cite{chiappa2018causal}.
\section{Edge Unfairness} \label{section4}
In this section, we quantify \textit{edge flow}, \textit{edge unfairness}, prove that eliminating edge unfairness eliminates cumulative unfairness, and quantify the \textit{potential to mitigate cumulative unfairness} when edge unfairness is reduced.

Edge flow along any edge, say $R \rightarrow J$ is the belief of observing a specific value of bail decision $J$ due to the influence of a specific value of race $R$ along $R \rightarrow J$. This can be extended to multiple direct edges from $\mathbf{M}=\{R, G\}$ to $X$.
\begin{definition}
Edge flow $\mathbb{P}^{\mathbf{M=m}}_{\text{flow}}(X=x)$ is defined as,
\begin{align}
\mathbb{P}^{\mathbf{m}}_{\text{flow}}(x) &= \frac{e^{\mathbb{\mathlarger{E}}_{\mathbf{m'} \sim \mathbb{P}(\mathbf{m'})}SE_{\boldsymbol{\pi},x}(\mathbf{m}, \mathbf{m'})}}{\underset{x}{\sum}e^{\mathbb{E}_{\mathbf{m'} \sim \mathbb{P}(\mathbf{m'})}SE_{\boldsymbol{\pi},x}(\mathbf{m}, \mathbf{m'})}}, ~~\text{where} ~\boldsymbol{\pi}=\{M \rightarrow X | M \in \mathbf{M}\}
\end{align} \label{edgeflow}
\end{definition}
Edge flow is formalized using direct effect
$SE_{\boldsymbol{\pi},x} (\mathbf{m}, \mathbf{m'})$
\citep{avin2005identifiability}. $SE_{\boldsymbol{\pi},x} (\mathbf{m}, \mathbf{m'})$ is the effect of $\mathbf{M=m}$ on $X=x$ along the direct edges from $\textbf{M}$ to $X$ irrespective of the value $\mathbf{m'}$ set along the indirect paths ensured by averaging. $SE_{\boldsymbol{\pi},x} (\mathbf{m}, \mathbf{m'})$ is identifiable because there is no recanting witness (see Definition 3, Theorem 1, and Theorem 2 in \cite{zhang2017causal}). A positive scaling like softmax ensures that the edge flow is a positive quantity which we use to prove Theorem \ref{boundtheorem}.

To quantify edge unfairness along $R \rightarrow J$, we first decompose $\mathbb{P}(J|Pa(J))$ into edge flows along the direct edges $\{X \rightarrow J | X \in Pa(J)\}$. The rationale is that the active trails $\{X \rightarrow J | X \in Pa(J)\}$ (Definition \ref{activetrail}) resulting from the dependencies $\{(J \not\perp X | Pa(J)\backslash X) | X \in Pa(J)\}$, which influence $J$'s value in $\mathbb{P}(J|Pa(J))$ (Theorem \ref{dseptheorem}), are same as the edges along which the edge flows from $Pa(J)$ to $J$ propagate. The theorem below formalizes this concept.

\begin{theorem}
The conditional probability distribution $\mathbb{P}(X=x|Pa(X)=pa(X))$ is a function $f^\mathbf{w}$ of $\text{FairFlow}_{x,pa(X)}$ and $\text{UnfairFlow}_{x,pa(X)}$
given by, \label{cptdecom}
\begin{align}
&\mathbb{P}(x|pa(X)) = f^\mathbf{w}(\text{FairFlow}_{x,pa(X)},\text{UnfairFlow}_{x,pa(X)}) \\
\text{where,} ~&~f^\mathbf{w}: \mathbb{W}^{|\mathbf{U}_{X}|+|\mathbf{F}_{X}|+1} \rightarrow [0, 1], \\ \text{subject to,} ~&\underset{x}{\sum} f^\mathbf{w}(\text{FairFlow}_{x,pa(X)},\text{UnfairFlow}_{x,pa(X)}) = 1, \label{probconst1}\\
&~~~f^\mathbf{w}(\text{FairFlow}_{x,pa(X)},\text{UnfairFlow}_{x,pa(X)}) \geq 0 \label{probconst2}
\end{align}
where $\mathbf{w}$ are the weights, $\mathbf{U}_X$ are the parents of $X$ along an unfair edge, $\mathbf{F}_X$ are the parents of $X$ along a fair edge, $\text{FairFlow}_{x,pa(X)}=\mathbb{P}^{\mathbf{F}_X=pa(X) _{\mathbf{F}_X}}_{\text{flow}}(X=x)$ and $\text{UnfairFlow}_{x,pa(X)}=\{\mathbb{P}^{A=pa(X)_A}_{\text{flow}}(X=x);~A \in \mathbf{U}_{X}\}$ 
\end{theorem}
\textit{\textbf{Proof:}} $\text{FairFlow}_{x,pa(X)}=\mathbb{P}^{pa(X)_{\mathbf{F}_X}}_{\text{flow}}(x)$ measures the effect of parents of X along the fair edges and $\text{UnfairFlow}_{x,pa(X)}= \{\mathbb{P}^{A=pa(X)_A}_{\text{flow}}(X=x);~A \in \mathbf{U}_{X}\}$ measures the effect of parents of X along the unfair edges [Definition \ref{edgeflow}]. Thus, $\underset{A \in \mathbf{U}_{X}} {\bigcup}\mathbb{P}^{pa(X)_A}_{\text{flow}}(x)$ and $\mathbb{P}^{pa(X)_{\mathbf{F}_X}}_{\text{flow}}(x)$ measure the effects along the direct edges $\{M \rightarrow X | M \in Pa(X)\}$. Further, the set of active trails resulting from the dependencies $\{(X \not\perp M | Pa(X)\backslash M) | M \in Pa(X)\}$, which influence $X$'s value in $\mathbb{P}(x|pa(X))$, is also $\{M \rightarrow X | M \in Pa(X)\}$ [Theorem \ref{dseptheorem}]. Hence, $\mathbb{P}(x|pa(X))$ can be formulated as a function of $\underset{A \in \mathbf{U}_{X}} {\bigcup}\mathbb{P}^{pa(X)_A}_ {\text{flow}}(x)$ and $\mathbb{P}^{pa(X)_{\mathbf{F}_X}}_ {\text{flow}}(x)$ provided the function satisfies the axioms of probability $\blacksquare$. \\


\begin{figure}[h!]
\centering
\includegraphics[height=6cm,width=0.65\linewidth]{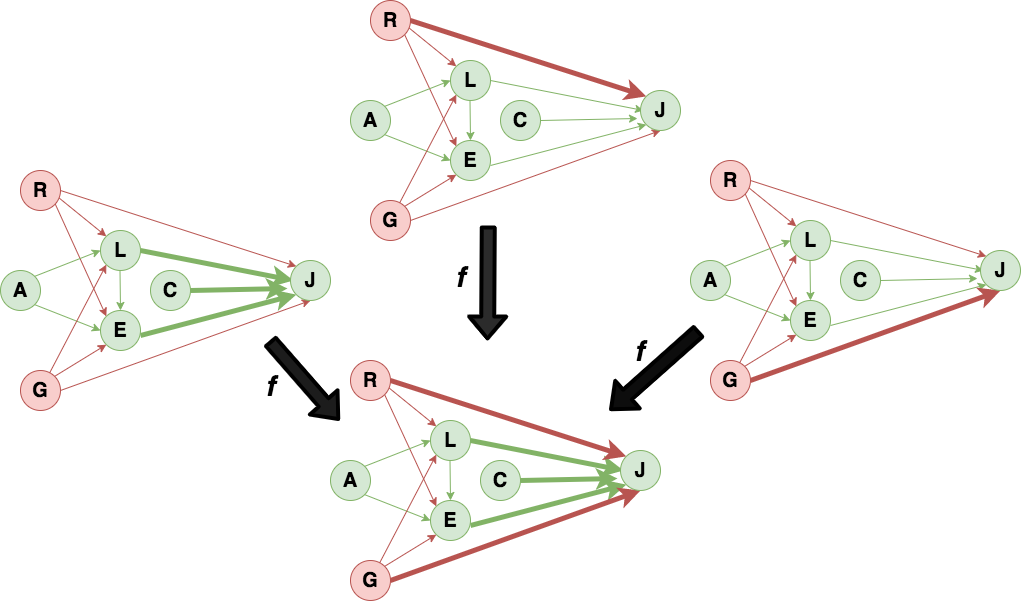}
\caption{Edge flows $\mathbb{P}^{\{L, C, E\}}_{\text{flow}}(J)$, $\mathbb{P}^{R}_{\text{flow}}(J)$ and $\mathbb{P}^{G}_ {\text{flow}}(J)$ interact via function $f$ to generate $\mathbb{P}(J|Pa(J))$. $\mathbb{P}^{\{L, C, E\}}_{\text{flow}}(J)$ result in effects along the fair edges $\{L \rightarrow J, E \rightarrow J, C \rightarrow J\}$ as shown in middle-left. $\mathbb{P}^{R}_{\text{flow}}(J)$ result in the effect along the unfair edge $\{R \rightarrow J\}$ as shown in top-center. $\mathbb{P}^{G}_{\text{flow}}(J)$ result in the effect along the unfair edge $\{G \rightarrow J\}$ as shown in the middle-right. The active trails resulting as a consequence of the dependencies in $\mathbb{P}(J|Pa(J))$ are $\{R \rightarrow J, L \rightarrow J, E \rightarrow J, C \rightarrow J, G \rightarrow J\}$ as shown in bottom.}
\label{modelgraph}
\end{figure}
This theorem aids in the formulation of edge unfairness in an unfair edge, say in $R \rightarrow J$ as the difference in $\mathbb{P} (J|Pa(J))$ with and without the edge flow in $R \rightarrow J$. \textit{Edge unfairness} is formalized below.

\begin{definition}
Edge unfairness $\mu_{e}$ of an unfair edge $e=K \rightarrow X$ is,
\begin{align}
&\mu_{e}=\mathbb{E}_{(x,pa(X)) \sim \mathbb{P}(x,pa(X))}\left[\frac{D^{K}_{x,pa(X)}}{\mathbb{P}^{pa(X)_K}_{\text{flow}}(x)}\right] \\ \label{edgeunfair}
\text{where,} ~&D^{K}_{x,pa(X)}=\bigg|\text{CPT}^{K}_{x,pa(X)}- \widetilde{\text{CPT}}^{K}_{x,pa(X)}\bigg|\\&\text{CPT}^{K}_{x,pa(X)}=f_X(\text{FairFlow}_{x,pa(X)}, \underset{A \in \mathbf{U}_{X}}{\bigcup}
\mathbb{P}^{pa(X)_A}_{\text{flow}}(x)) \\&\widetilde{\text{CPT}}^{K}_{x,pa(X)}=f_{X}(\text{FairFlow}_{x,pa(X)},\underset{A \in \mathbf{U}_{X}\backslash K}{\bigcup}\mathbb{P}^{pa(X)_A} _{\text{flow}}(x),\mathbb{P}^{pa(X)_K}_{\text{flow}}(x)=0)
\end{align}
\end{definition}
Edge unfairness $\mu_{e}$ is the unit contribution of edge flow $\mathbb{P}^{K}_{\text{flow}}(X)$ to $\mathbb{P}(X|Pa(X))$. $\mu_{e}$ measures the difference in $\mathbb{P}(X | Pa(X))$ with the edge flow along $e$, given by $\text{CPT}^{K}_{x,pa(X)}$, and without the edge flow along $e$, given by $\widetilde{\text{CPT}}^{K} _{x,pa(X)}$. We measure per unit edge flow to capture that a large $D_{X,K}$ compared to $\mathbb{P}^{K}_ {\text{flow}}(X)$ still results in large $\mu_{e}$ even though $D^{K}_{X,Pa(X)}$ is small. \\

Now, we quantify cumulative unfairness. The objective for introducing cumulative (overall) unfairness is twofold: (1)To prove that eliminating edge (local) unfairness eliminates cumulative unfairness (2)To formulate the potential to mitigate cumulative unfairness. We combine direct and indirect discrimination, discussed in Section 3 of \cite{zhang2017causal}, to define cumulative unfairness towards sensitive nodes $\mathbf{S=s}$ in decision $Y=y$. Cumulative unfairness $C_{\mathbf{S=s},Y=y}$ is, 
\begin{align}
    C_{\mathbf{S=s},Y=y} = \mathbb{E}_{\mathbf{s'}\sim\mathbb{P}(\mathbf{s'})} TE_{Y=y}(\mathbf{s, s'})
\end{align}
$C_{\mathbf{S=s},Y=y}$ measures the impact on outcome $Y=y$ when $\mathbf{S}$ is forcibly set to $\mathbf{s}$ along the unfair paths from $\mathbf{S}$ to $Y$ irrespective of the value set along other paths. Since all edges emanating from a sensitive node are potential sources of unfairness, the total causal effect $TE$ (see Definition \ref{te}) is used to formulate cumulative unfairness. Proving the result that eliminating edge unfairness $\mu_e$ in all unfair edges $e$ eliminates cumulative unfairness $C_{\mathbf{S=s},Y=y}$ is not straightforward as $C_{\mathbf{S=s},Y=y}$ cannot be expressed in terms $\mu_e$. We first upper bound $|C_{\mathbf{S=s},Y=y}|$ by $C^{\text{upper}}_{\mathbf{S=s},Y=y}$ that can be expressed in terms of $\mu_e$. Then, the result follows from the theorem.
\begin{theorem}
The magnitude of cumulative unfairness in decision $Y=y$ towards sensitive nodes $S=s$, $C_{\mathbf{S=s},Y=y}$, is upper bounded by $C^{\text{upper}}_{\mathbf{S=s},Y=y}$ as shown below,
\begin{align}
|C_{\mathbf{S=s}, Y=y}| \leq C^{\text{upper}}_{\mathbf{S=s},Y=y}
\end{align} \label{boundtheorem}
where,

\begin{align}
&C^{\text{upper}}_{\mathbf{S=s},Y=y}=\underset{\mathbf{s'}\in\mathbf{S}\backslash\mathbf{s}}{\sum}~\mathbb{P}(\mathbf{s'})\left[\underset{v\in\mathbf{V}\backslash \{\textbf{S},Y\}}{\sum} ~\underset{V\in\mathbf{V}\backslash \{\mathbf{S},Y\},Y=y} {\prod} \underset{~A \in \mathbf{U}_{V}} {\sum}\left[\frac{\mathbb{P}^{\mathbf{s}_A \lor pa(V)_A}_{\text{flow}} (v)}{\mathbb{P}(v,pa(V))|_{\mathbf{s}}}+\frac{\mathbb{P}^{\mathbf{s'}_A \lor pa(V)_A}_{\text{flow}}(v)}{\mathbb{P}(v, pa(V))|_ {\mathbf{s'}}}\right] \mu_{A \rightarrow V}\right] \label{cumuppb2} \\ 
&~~~~~~~~~~~~~~~~~~~~~
\mathbf{s}_A \lor pa(V)_A \equiv A = 
\begin{dcases}
\mathbf{s}_A, & ~\text{if} ~A \in \textbf{S} \\[1ex]
pa(V)_A & ~\text{otherwise}
\end{dcases} \label{notationor}
\end{align} \label{theoremcum}
\end{theorem}
\textbf{Proof Sketch of Theorem:}\\
Cumulative unfairness $C_{\mathbf{S=s}, Y=y}$ cannot be expressed in terms of edge unfairness when the conditional probability is modeled by a non-parametric model $f$. Therefore, we write $C_{\mathbf{S=s},Y=y}$ in terms of conditional probabilities $CPT$s. Each $CPT$ is substituted by its functional model $f$ (see Theorem \ref{cptdecom}), because the edge unfairness is expressed in terms of $f$ (see Definition \ref{edgeunfair}). To bring edge unfairness $\mu_e$ into the formulation, we upper bound each $f$ of a node present along an unfair edge $e$ with the following quantities: edge unfairness $\mu_e$ and $f$ having no edge flow along $e$. The rationale of this step comes from the definition of edge unfairness $\mu_e$ (see Definition \ref{edgeunfair}) and the fact that the modulus operation is a non-negative quantity. This proves the result $C_{\mathbf{S=s}, Y=y} \leq C^{\text{upper}}_{\mathbf{S=s},Y=y}$. By following similar steps and using modulus operation to lower bound $f$, we arrive at $C_{\mathbf{S=s}, Y=y} \geq -C^{\text{upper}}_ {\mathbf{S=s},Y=y}$. \textit{(see Supplementary for full proof)} $\blacksquare$
\begin{corollary}
The cumulative unfairness in decision $Y=y$ towards sensitive nodes $S=s$, $C_{\mathbf{S=s},Y=y}$, is non-existent when edge unfairness $\mu_{e}$ in all unfair edges is eliminated. \label{boundtheoremcorol}
\end{corollary}
We now measure the potential to mitigate cumulative unfairness when edge unfairness is reduced. Using the \textit{potential} measure and \textit{edge unfairness}, agencies can then prioritize the unfair edges before taking measures to mitigate them. \\

\textit{Sensitivity} measures the variation in $C^{\text{upper}}_{\mathbf{S=s},Y=y}$ when edge unfairness $\mu_{e}$ in unfair edge $e$ is varied. Since $C^{\text{upper}}_{\mathbf{S=s},Y=y}$ is a linear function in edge unfairness, higher order derivatives ($\geq 2$) of $C^{\text{upper}}_{\mathbf{S=s},Y=y}$ with respect to $\mu_{e}$ are 0.
\begin{definition}
\textbf{Sensitivity} of $C^{\text{upper}}_{\mathbf{S=s},Y=y}$ w.r.t edge unfairness in edge $e$ $\mu_{e}$ is,
\begin{align}
&S^{\mathbf{S=s},Y=y}_{e}=\frac{\partial C^{\text{upper}}_{\mathbf{S=s},Y=y}}{\partial \mu_{e}}\bigg|_{\boldsymbol{\mu}^*} \label{sens}
\end{align} 
where, $\boldsymbol{\mu}^{*}$ are the current edge unfairness obtained from observational distribution $\mathbb{P}(\mathbf{V})$.
\end{definition}
The following quantity measures the potential to mitigate $C^{\text{upper}}_{\mathbf{S=s},Y=y}$ when edge unfairness $\mu_{e}$ in unfair edge $e$ is decreased.  $C^{\text{upper}}_ {\mathbf{S=s},Y=y}$ is used to measure the potential contrary to $C_{\mathbf{S=s},Y=y}$ because $C_{\mathbf{S=s},Y=y}$ cannot be expressed in terms of edge unfairness. Experiment \ref{relationship_theorem3} validates that decreasing $C^{\text{upper}}_ {\mathbf{S=s},Y=y}$ decreases $C_{\mathbf{S=s},Y=y}$. 
\begin{definition}
\textbf{Potential to Mitigate Cumulative Unfairness} when $\mu_{e}$ is decreased. 
\begin{align}
P^{\mathbf{S=s},Y=y}_{e} =
\begin{dcases}
-\bigg|S^{\mathbf{S=s},Y=y}_{e}\bigg| & \text{if }C^{\text{upper}}_{\mathbf{S=s},Y=y}=0 \\[1ex]
S^{\mathbf{S=s},Y=y}_{e} & \text{if } C^{\text{upper}}_{\mathbf{S=s},Y=y} > 0
\end{dcases} 
\end{align} \label{potential}
\end{definition}
$P^{\mathbf{S=s},Y=y}_{e}$ states that if $C^{\text{upper}}_{\mathbf{S=s},Y=y}=0$, then $C^{\text{upper}}_{\mathbf{S=s},Y=y}$ deviates from 0 (indicative of no-discrimination) as edge unfairness is decreased. The potential of $C^{\text{upper}}_{\mathbf{S=s},Y=y}$ to move towards 0 or to get mitigated is then quantified by $-|S^{\mathbf{S=s},Y=y}_{e}|$ wherein negative is due to $C^{\text{upper}}_{\mathbf{S=s},Y=y}$ deviating from 0. Similarly, one can analyze the other case.

\section{Unfair Edge Prioritization \& Discrimination Removal} \label{algorithms}
Based on the theorems and the definitions, we present pseudo-codes for fitting the $CPTs$ in Algorithm \ref{fitCPT}, computing priority of the unfair edges in Algorithm \ref{priority}, and removing discrimination in Algorithm \ref{disc_algo}. Algorithm \ref{priority} aids the agencies to mitigate unfairness underlying the unfair edges in the real-world during the data generation phase. Algorithm \ref{disc_algo} de-biases data distribution after the data generation phase. Algorithm \ref{priority} calls Algorithm \ref{fitCPT}. Algorithm \ref{fitCPT} does not call Algorithm \ref{priority} and Algorithm \ref{disc_algo}. Algorithm \ref{disc_algo} does not call Algorithm \ref{priority} and Algorithm \ref{fitCPT}.  \\

\noindent (1)\textbf{fitCPT() Algorithm \ref{fitCPT}}: It takes the causal model $(\mathbb{G},\mathbb{P})$, the set of unfair edges $\mathbf{E}^{\text{unfair}}_{\mathbb{G}}$, and the attribute $X$ as inputs and approximates the $CPTs$ $\mathbb{P}(X|Pa(X))$ by the model $f^\mathbf{w}$ using the least-squares loss.
\begin{algorithm}[H]
    \caption{fitCPT($\mathbb{G}$,$\mathbb{P}$,$\mathbf{E}^{\text{unfair}}_{\mathbb{G}}$,X)}
    \begin{algorithmic}[1]
    \State \text{Initialize} $\mathbf{w}$ \text{randomly}
    \State $\mathbf{Y} \leftarrow \mathbb{P}(X=x|Pa(X)=pa(X))$
    \State $\text{Compute}~\text{FairFlow}_{x,pa(X)}$ \text{and} $\text{UnfairFlow}_{x,pa(X)} \hfill \text{(Theorem \ref{cptdecom})}$
    \State $\hat{\mathbf{Y}}_X(\mathbf{w}) \leftarrow f^\mathbf{w}(\text{FairFlow}_{x,pa(X)}, \text{UnfairFlow}_{x,pa(X)}) \hfill \text{(Theorem \ref{cptdecom})}$
    \State $\mathbf{w}^* \leftarrow \text{arg}\min_{\mathbf{w}}||\mathbf{Y} - \hat{\mathbf{Y}}_X(\mathbf{w})||^2$ \text{subject to} Eq. \ref{probconst1} \& Eq. \ref{probconst2} \\
    \textbf{Output:}~$\mathbf{w}^*$
    \end{algorithmic} \label{fitCPT}
\end{algorithm}

\noindent (2)\textbf{computePriority() Algorithm \ref{priority}}: It computes priorities of the unfair edges based on the edge unfairness and the potential to mitigate the cumulative unfairness. The priorities can be used to address unfairness in the real world.
\begin{algorithm}[H]
    \caption{computePriority($\mathbb{G},\mathbb{P},\mathbf{E}^{\text{unfair}}_{\mathbb{G}}$,$\mathbf{s}$,$y$,$w_u$,$w_{p}$)}
    \begin{algorithmic}[1]
    \State $\mathbf{w}^{*} = \{\}$ \Comment{Optimal weights of the approximated \textit{CPT}s}\\
    \text{for} $V$ \text{in} $\mathbf{V}$ \text{do}
        \State ~~~$\mathbf{w}^*_V \leftarrow \text{fitCPT}(\mathbb{G},\mathbb{P},\mathbf{E}^{\text{unfair}}_{\mathbb{G}},V)$ \Comment{See Algorithm \ref{fitCPT}}
        \State ~~~$\mathbf{w}^{*} \leftarrow \mathbf{w}^{*}\cup \{\mathbf{w}^*_V\}$
    \State priorityList = $\{\}$\\
    \text{for} $e=S \rightarrow V ~\text{in} ~\mathbf{E}^{\text{unfair}}_{\mathbb{G}}$
    \State ~~~Compute $\mu_{e}$ using $f^{\mathbf{w}^{*}_{V}}$ from Eq. \ref{edgeunfair}
    \State ~~~Compute $P^{\mathbf{S=s},Y=y}_{e}$ \Comment{See Definition \ref{potential}}
    \State \text{priority} = $w_u{\mu}_{e} + w_{p} P^{\mathbf{S=s},Y=y}_{e}$ \Comment{$w_u$ and $w_p$ are weights}
    \State \text{priorityList = priorityList} $\cup \{(e,\text{priority)}\}$\\
    \textbf{Output:} $\text{priorityList}$
    \end{algorithmic} \label{priority}
\end{algorithm}

\noindent (3)\textbf{removeDiscrimination() Algorithm \ref{disc_algo}}: It removes discrimination by regenerating new \textit{CPT}s for the causal model $(\mathbb{G},\mathbb{P})$ with unfair edges $\mathbf{E}^{\text{unfair}}_{\mathbb{G}}$.  These \textit{CPT}s are approximated by solving an optimization problem of minimizing the overall edge unfairness subject to the axioms of probability as constraints. A \textit{data utility} term, which is the Mean Squared Error (\textit{MSE}) between $\mathbb{P}(\mathbf{V})$ and the new joint distribution computed from the product of approximated \textit{CPT}s, is added to the objective function to ensure that the influences from other insensitive nodes are preserved. For instance, a sensitive node like religious belief $R$ can have insensitive nodes like literacy $L$ as a parent. By minimizing only the edge unfairness in the objective function, indirect influences like $L \rightarrow R \rightarrow J$ can get altered, thereby not preserving data utility. Also, this algorithm gets away with the subjectively chosen threshold of discrimination in the constraints, unlike previous works. This circumvents the problem of the regenerated data distribution being unfair had a smaller threshold been chosen.
\begin{algorithm}[H]
\caption{removeDiscrimination($\mathbb{G}$,$\mathbb{P}$,$\mathbf{E}^{\text{unfair}}_{\mathbb{G}}$)}
\begin{algorithmic}[1]
\State $\mathbf{w^*} \leftarrow \text{argmin}_{\mathbf{w}} \underset{e \in \mathbf{E}^{\text{unfair}}_{\mathbb{G}}}{\mathlarger{\sum}} \mu_{e} + \left\lVert \mathbb{P}(\mathbf{V}) - \underset{Z \in \mathbf{V}} {\mathlarger{\prod}} 
f^\mathbf{w}(\text{FairFlow}_{z,pa(Z)},\text{UnfairFlow}_{z,pa(Z)})
\right\rVert^2~$ 
\NoNumber{\hfill $\text{subject to} ~\text{Eq.}~\ref{probconst1}~\text{and}~\text{Eq.}~\ref{probconst2}$ (Theorem \ref{cptdecom})}
\State $\mathbb{P}_{\text{new}}(\mathbf{V}) \leftarrow \underset{Z\in \mathbf{V}}{\mathlarger{\prod}} 
f^\mathbf{w^{*}}(\text{FairFlow}_{z,pa(Z)},\text{UnfairFlow}_{z,pa(Z)})$ \hfill \text{(Theorem \ref{cptdecom})}\\
$\textbf{Output:}~\mathbb{P}_{\text{new}}(\mathbf{V})$
\end{algorithmic} \label{disc_algo}
\end{algorithm}

\section{EXPERIMENTS}
\label{section:exper}
In this section, we perform experiments to validate the model and input specifications for approximating \textit{CPT}s using the criminal recidivism graph as shown in Fig. \ref{main_graph}. The effectiveness and efficiency of algorithms depend upon their building blocks: (1) Edge Unfairness, (2) Theorem 3, (3) Algorithm 1. Hence, we focus the experimental section on \textit{Edge Unfairness} formulation, Theorem \ref{boundtheorem} and Algorithm \ref{fitCPT} as they are used in Algorithm \ref{priority} and Algorithm \ref{disc_algo}. In particular, we analyze the relationship between \textit{Cumulative Unfairness} and the upper bound of \textit{Cumulative Unfairness}, and the applicability of our method to realistic scenarios where the causal model and the \textit{CPT}s are unavailable. We first define the causal model by constructing the \textit{CPT}s. 

\subsection{Causal Model}\label{causal_exp}
This paper uses the causal graph shown in Fig. \ref{main_graph} for experiments. This graph is similar to the one constructed in \cite{vanderweele2011causal}; the difference is in the usage of the defendant's attributes like race as nodes contrary to the judge's attributes. The values taken by the nodes are discrete and specified in Supplementary material. For each attribute $V$ in the graph, the conditional probability distribution $\mathbb{P}(V|Pa(V))$ is generated by the following quantities: 
\begin{itemize}
    \item \textit{Parameters}: $\theta_{A\rightarrow V} \in [0, 1] ~ \forall A \in Pa(V)$ where $\theta_{A \rightarrow V}$ quantifies the direct influence of parent $A$ on $V$ that is independent of the specific values taken by $A$ and $V$. $\theta_{A \rightarrow V}$ is a property of the edge $A \rightarrow V$.
    \item \textit{Scores}: $\lambda_{A=a\rightarrow V=v} \in [0,1] ~\forall A \in Pa(V)$ where $\lambda_{A=a\rightarrow V=v}$ quantifies the direct influence of parent $A$ on $V$. It is dependent on the specific values of $A$ and $V$.\\
\end{itemize}
\textit{CPT} of node $V$ is computed as the weighted sum of $\lambda_{A=a\rightarrow V=v}$ with $\theta_{A\rightarrow V}$ being the weights,
\begin{align}
\mathbb{P}(v | pa(V)) = \underset{A \in Pa(V)}{\sum} \theta_{A\rightarrow V}\lambda_{A=pa(V)_A\rightarrow V=v} \label{modeleq}
\end{align}
To ensure that the \textit{CPT}s satisfy marginality conditions, the following constraints are defined over the parameters and scores: $\underset{A\in Pa(V)}{\sum} \theta_{A\rightarrow V} = 1$ and $\underset{v}{\sum} \lambda_{A=a\rightarrow V=v} = 1, \forall A\in Pa(V)$. We generate $625$ models with different combinations of $\{\theta_{A\rightarrow J}, \theta_{B \rightarrow E} | A \in Pa(J), B \in Pa(E)\}$ that are used to generate \textit{CPT}s while keeping $\lambda_{A=a\rightarrow V=v}$ fixed.

\subsection{Approximating the CPTs}\label{cpt_exp}
We implement Algorithm \ref{fitCPT} for each \textit{CPT} and solve the constrained least-squares problem (CLSP) to find the optimal solution $\mathbf{w}^*$ (Algorithm \ref{fitCPT}: Step 5). CLSP is a well-known optimization problem for the linear model. In the case of a non-linear model, we implement a neural network and apply Adam optimizer \citep{kingma2014adam} with default hyper-parameters to minimize the \textit{MSE} loss. These two models were implemented using \textsc{scikit-learn} library and \textsc{PyTorch} library respectively \citep{NEURIPS2019_9015}. 

\subsection{Experiment 1: Relationship between $|C_{\mathbf{S=s},Y=y}|$ and $C^{\text{upper}}_{\mathbf{S=s},Y=y}$} \label{relationship_theorem3}
~~~~\textit{\textbf{Utility:}}
We know from Theorem \ref{boundtheorem} that $C^{\text{upper}}_ {\mathbf{S=s},Y=y}=0$ and $C_{\mathbf{S=s},Y=y}=0$ when edge unfairness $\mu_e=0$ in all the unfair edges $e$. Here, we investigate whether decreasing $C^{\text{upper}}_{\mathbf{S=s},Y=y}$ decreases $|C_{\mathbf{S=s},Y=y}|$. This investigation provides utility for the formulation of \textit{Potential to Mitigate Cumulative Unfairness} quantity in terms of $C^{\text{upper}}_{\mathbf{S=s},Y=y}$ as $C_{\mathbf{S=s},Y=y}$ cannot be expressed in terms of edge unfairness.

\begin{figure}[h!]
\centering
\subfigure[]{\includegraphics[width=0.47 \linewidth] {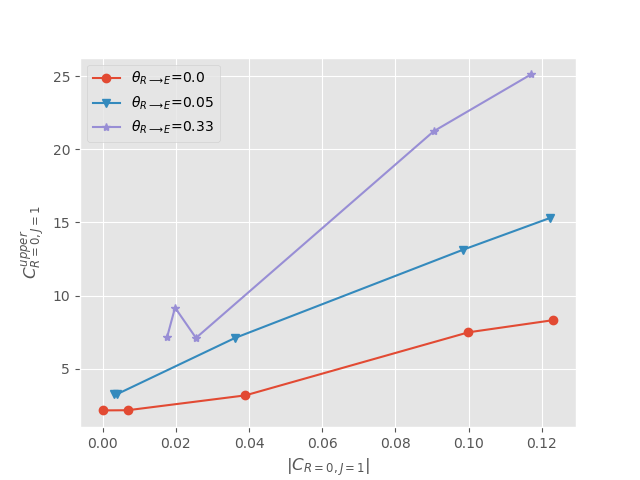}}
\subfigure[]{\includegraphics[width=0.47\linewidth] {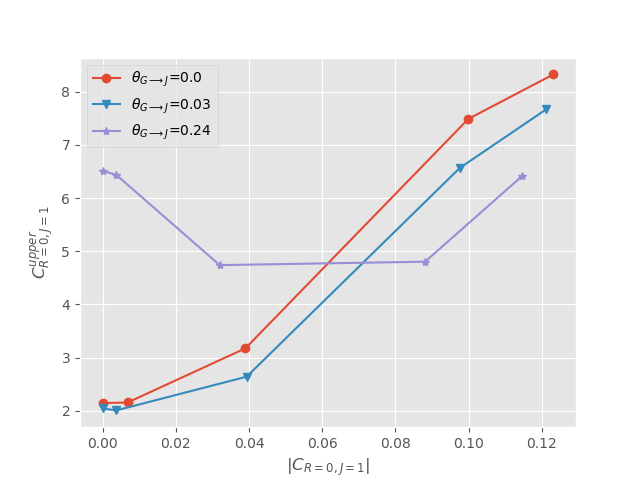}}
\caption{Decreasing $C^{\text{upper}}_{\mathbf{S=s},Y=y}$ decreases $|C_{\mathbf{S=s},Y=y}|$ when edge unfairness $\mu_e$ in all unfair edges $e$ are small. (a) $\theta_{R\rightarrow J}$, indicative of edge unfairness $\mu_{R\rightarrow J}$, is varied for different $\theta_{R\rightarrow E}$. (b) $\theta_{R\rightarrow J}$ is varied for different $\theta_{G\rightarrow J}$. When $\theta_{R\rightarrow J}$ is large, decreasing $C^{\text{upper}}_ {\mathbf{S=s},Y=y}$ does not decrease $|C_{\mathbf{S=s},Y=y}|$ as seen from the purple graph.}
\label{cum_uppexp}
\end{figure}

\textit{\textbf{Setting:}}
We set $\theta_{A\rightarrow V}=0$ for all $A$ along an unfair edge to $0$ except the racial parent $R$. $\theta_{A\rightarrow V}$ is indicative of edge unfairness $\mu_{A\rightarrow V}$ from Experiment \ref{cpt_exp}. Next, we plot $C^{\text{upper}}_{R=0,J=1}$ and $|C_{R=0,J=1}|$ with varying $\theta_{R\rightarrow J}$ for different values of $\theta_{R\rightarrow E}$ (Fig. \ref{cum_uppexp}(a)) and $\theta_{G\rightarrow J}$ (Fig. \ref{cum_uppexp}(b)) respectively.

\textit{\textbf{Inference:}}
We observe that for small edge unfairness, decreasing $C^{\text{upper}}_{R=0,J=1}$ decreases $|C_{R=0,J=1}|$. Both converge to $0$ when all the edge unfairness are eliminated. This inference helps the policy makers to mitigate cumulative unfairness $|C_{R=0,J=1}|$ by mitigating the upper bound $C^{\text{upper}}_{R=0,J=1}$. On the other hand, as we increase the edge unfairness in the edges other than $R\rightarrow J$, this linear trend diminishes as observed for $\theta_{R\rightarrow E}=0.33$ (Fig. \ref{cum_uppexp}(a)) and $\theta_{G\rightarrow J}=0.24$ (Fig. \ref{cum_uppexp}(b)). 


\subsection{Experiment 2: Edge Unfairness with Finite data}
~~~~\textbf{\textit{Utility:}} We investigate the applicability of our approach to realistic scenarios where the causal model and the \textit{CPT}s are unavailable. We do not dwell on discovering causal structures using finite data.\footnote{TETRAD software discussed in \cite{ramsey2018tetrad} can be used for this purpose.} 
Instead, we focus on estimating \textit{CPT}s using a finite amount of data and compare the edge unfairness calculated using original \textit{CPT}s, $\mathbb{P}$, with the one by estimated \textit{CPT}s, $\mathbb{P}^m$, where $m$ is the number of samples drawn randomly from $\mathbb{P}$ for estimation. Intuitively, the distance should decrease as $m$ increases because a large number of i.i.d. samples produce a better approximation of the original distribution $\mathbb{P}$, thereby reducing the euclidean distance. 

\begin{figure}[h!]
\centering
\subfigure[]{\includegraphics[width=0.43\linewidth]{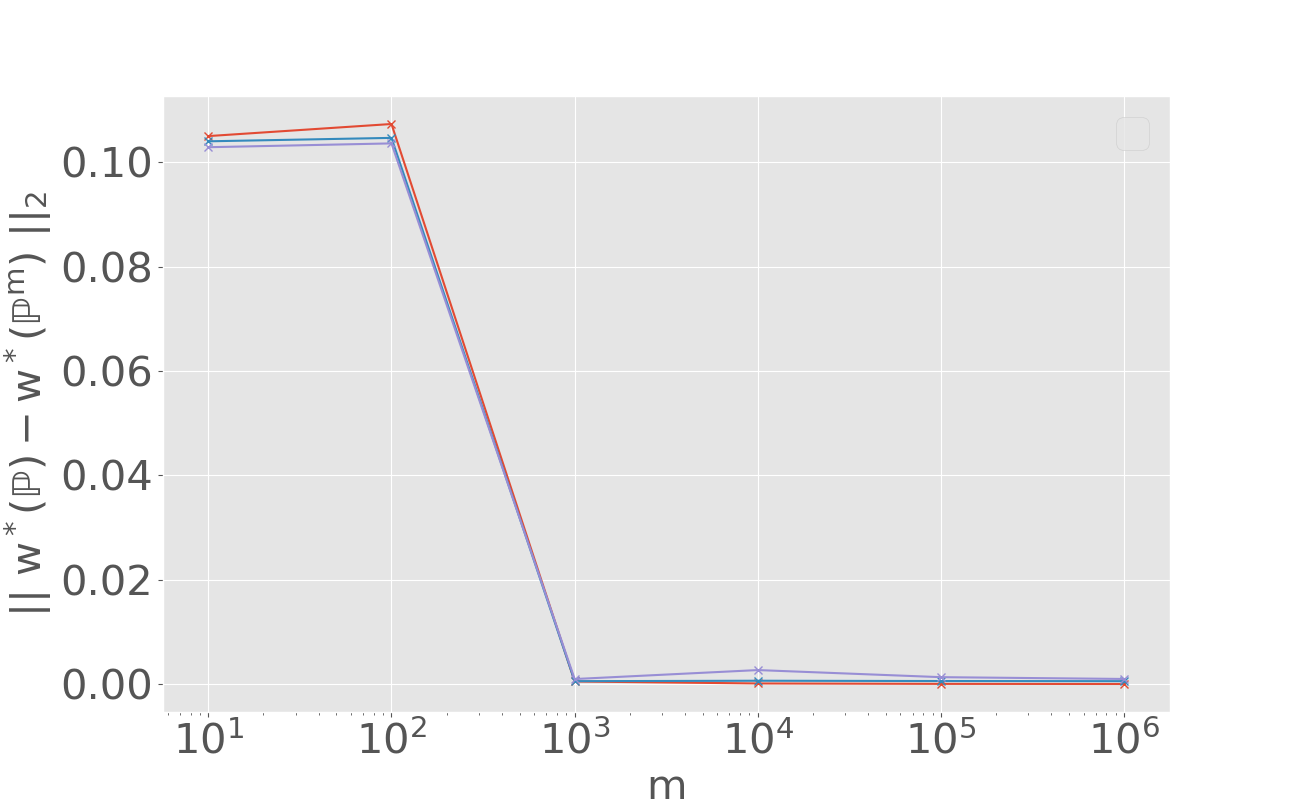}}
\subfigure[]{\includegraphics[width=0.43\linewidth]{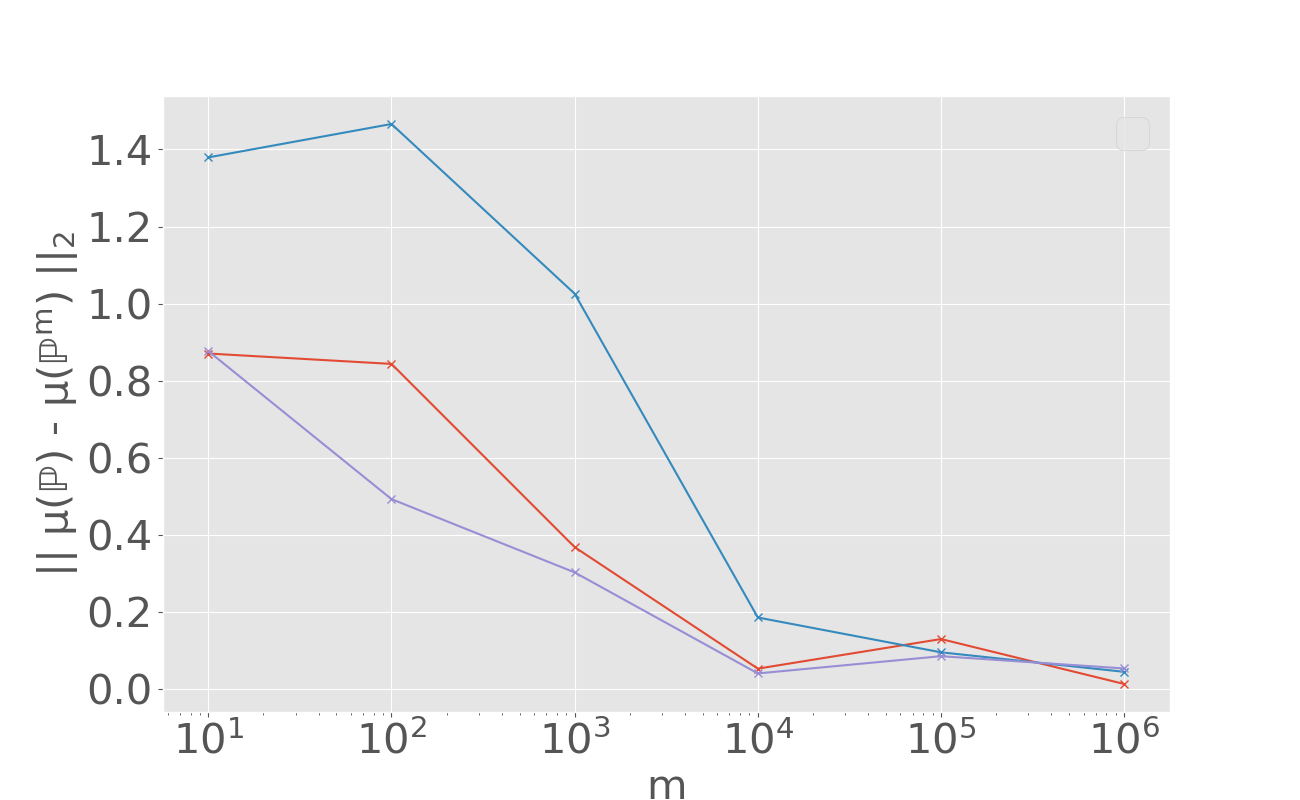}} 
\caption{Edge unfairness decreases as the number of samples increases. (a) $||\mathbf{w}^*(\mathbb{P}) - \mathbf{w}^*(\mathbb{P}^m)||_2=D_{L}^{\mathbb{P}}(m)$ vs. number of samples $m$. (b) $||\mu(\mathbb{P}) - \mu(\mathbb{P}^m)||_2= D_{NL}^{\mathbb{P}}(m)$ vs. number of samples $m$.} \label{finite}
\end{figure}

\textbf{\textit{Setting:}} In Fig. \ref{finite}, we plot the euclidean distance $D_{L}^{\mathbb{P}}(m)$ between $\mathbf{w}^*(\mathbb{P})$ and $\mathbf{w}^*(\mathbb{P}^m)$ by varying $m$. Here, \textit{CPT}s are approximated using the linear model for different distributions $\mathbb{P}$ that are randomly generated as shown in different colors. Similarly, the euclidean distance $D_{NL}^{\mathbb{P}}(m)$ between $\mu(\mathbb{P})$ and $\mu(\mathbb{P}^m)$ assuming the non-linear model is shown in Fig. \ref{finite}(b). 

\textbf{\textit{Inference:}} We observe that $\mathbf{w}^*(\mathbb{P}^m)$ moves closer to $\mathbf{w}^*(\mathbb{P})$ as $m$ increases. Moreover, since $\mathbb{P}$ was randomly generated, we also observe that there exists an empirical bound over Euclidean distance for a given $m$. For instance, in Fig. \ref{finite}(a), $D_{L}^{\mathbb{P}}(m)$ is less than $0.01$ for $m$ greater than $10^3$. A similar observation can be made in Fig. \ref{finite}(b). Further, more samples are required to make $D_{L}^{\mathbb{P}}(m)$ and $D_{NL}^{\mathbb{P}}(m)$ comparable. For instance, around $10^3$ samples are required to observe $D_{L}^{\mathbb{P}}(m)=0.01$, while at least $10^4$ samples are required to observe $D_{NL}^{\mathbb{P}}(m) \approx 0.2$. The presence of an empirical bound motivates one to investigate the possibility of a theoretical bound over the Euclidean distance. In addition to the above experiments, we empirically show that the \textit{Edge Unfairness} is a property of an edge and discuss the benefits of using a non-linear model in the Supplementary Material.

\section{Related Work}
~~~~~\textbf{Mitigating Unfairness in the Data Generation Phase:} \cite{gebru2018datasheets} suggests documenting the dataset by recording the motivation and creation procedure. However, it does not attempt to provide a solution for mitigation with limited resources. 

\textbf{Assumptions:}  \cite{zhang2017causal} assumes that the sensitive variable $S$ has no parents as it is an inherent nature of the individual. We follow \cite{zhang2019boundscausal} that relaxes this assumption because sensitive nodes such as religious belief can have parents like literacy $L$. \cite{nabi2018fair} and \cite{chiappa2019path} propose discrimination removal procedures in the continuous node setting by handling the non-identifiability issues. We do not discuss the continuous variable setting to avoid digressing into the intractability issues. \cite{wu2019pc} formulates cumulative unfairness as a solution to the optimization problem for the semi-markovian setting. We restrict our discussion to the markovian setting to avoid digressing into the challenges of formulating cumulative unfairness in terms of edge unfairness. \cite{ravishankar2020causal} solves the problem for the trivial linear case when the cumulative unfairness can be expressed in terms of edge unfairness. 

\textbf{Edge Flow:} Decomposing direct parental dependencies of a child into independent contributions from each of its parents helps in quantifying the edge flow. \cite{srinivas1993generalization}, \cite{kim1983computational}, and \cite{henrion2013practical} separate the independent contributions by using unobserved nodes in the representation of causal independence. To overcome the issues of intractability in unobserved nodes, \cite{heckerman1993causal} proposed a temporal definition of causal independence. It states that if and only cause $c$ transitions from time $t$ to $t+1$, then the effect's distribution at time $t+1$ depends only on the effect and the cause at time $t$, and the cause at time $t+1$. Based on this definition, a belief network representation is constructed with the observed nodes that make the probability assessment and inference tractable. \cite{heckerman1994new} proposes a temporal equivalent of the temporal definition of \cite{heckerman1993causal}. The aforementioned works do not quantify the direct dependencies from the parents onto the child as in our work.

\textbf{Edge Unfairness:} Multiple statistical criteria have been proposed to identify discrimination \citep{berk2018fairness} but it is mathematically incompatible to satisfy them all when base rates of the dependent variable differ across groups \citep{chouldechova2017fair, kleinberg2016inherent}. Consequently, there is an additional task of selecting which criterion has to be achieved. Moreover, statistical criteria caution about discrimination but do not help in identifying the sources of unfairness. \cite{zhang2017causal} uses path-specific effects to identify direct and indirect discrimination after data is generated but does not address the problem of mitigating unfairness in the data generation phase. Unlike \cite{zhang2017causal} that uses the presence of a redlining attribute in an indirect path and the presence of a sensitive node on a direct path to determine the unfairness of a path, our work uses the notion of an unfair edge as the potential source of unfairness akin to \cite{chiappa2018causal}.

\textbf{Discrimination Removal Procedure:} \cite{zhang2017causal} and \cite{kusner2017counterfactual} remove discrimination by altering the data distribution. \textit{Firstly}, the optimization technique in \cite{zhang2017causal} and the sampling procedure in \cite{kusner2017counterfactual}, scale exponentially in the number of nodes (and values taken by the sensitive nodes) that eventually increases the time to solve the quadratic programming problem. \textit{Secondly}, the constraints in \cite{zhang2017causal} depend on a subjectively chosen threshold of discrimination that is disadvantageous because the regenerated data distribution would remain unfair had a smaller threshold been chosen. Our paper formulates a discrimination removal procedure without exponentially growing constraints and a threshold of discrimination.

\section{CONCLUSION}
\label{conclusion}
We introduce the problem of quantifying edge unfairness in an unfair edge. We give a novel formulation that models $CPTs$ in terms of edge flows to quantify edge unfairness. We prove a result that eliminating edge unfairness eliminates cumulative unfairness. Proving this result is not straightforward because cumulative unfairness cannot be expressed in terms of edge unfairness when $CPTs$ are modeled as a non-parametric function of the edge flows. Hence, we prove the result via an intermediate theorem that upper bounds the magnitude of cumulative unfairness by a quantity that can be expressed in terms of edge unfairness. To analyze the impact of edge unfairness on cumulative unfairness, we quantify the potential to mitigate cumulative unfairness when edge unfairness is decreased. This formulation uses the upper bound of cumulative unfairness as it can be expressed in terms of edge unfairness. Experimental results validate that mitigating cumulative unfairness mitigates its upper bound as well, thereby establishing the rationale for using the upper bound of cumulative unfairness in the formulation. Using the theorem result and measures, we present an unfair edge prioritization algorithm and a discrimination removal algorithm. The unfair edge prioritization algorithm gives tangible directions to agencies to mitigate unfairness in the real world while the data is being generated. There is no utility in making cautionary claims of potential discrimination when it is not complemented with information that aids in mitigating unfairness causing discrimination. On the other hand, the discrimination removal algorithm de-biases data after the data is generated. In the future, we aim to evaluate the impact of edge unfairness on subsequent stages of the machine learning pipeline such as selection, classification, etc. We also plan to extend to the semi-Markovian causal model \citep{wu2019pc} and continuous nodes settings \citep{nabi2018fair}.


\bibliography{acml21}

\newpage
\noindent \textbf{Title:} A Causal Approach for Unfair Edge Prioritization and Discrimination Removal
\section*{Supplementary Material}
\subsection*{1. Choices for $f^\mathbf{w}$}
\label{sec:choicesf}
We present two instances for $f^\mathbf{w}$. The list is not limited to these and can be extended as long as $f^\mathbf{w}$ satisfies the constraints of the conditional probability (Eq. \ref{probconst1}, \ref{probconst2}).
\begin{enumerate}[topsep=0pt]
    \item $f^\mathbf{w}$ is a linear combination in the inputs where,
    \begin{align}
    &\resizebox{0.85\textwidth}{!}{$f^\mathbf{w}(\mathbb{P}^{pa(X)_{\mathbf{F}_X}}_{\text{flow}}(X), \underset{pa(X)_A \in \mathbf{U}_{X}}{\bigcup}\mathbb{P}^{pa(X)_A}_ {\text{flow}}(X))=w_{\mathbf{F}_X \rightarrow X}\mathbb{P}^{pa(X)_{\mathbf{F}_X}}_{\text{flow}}(X) + \underset{A \in \mathbf{U}_{X}}{\sum}w_{A \rightarrow X}\mathbb{P}^{pa(X)_A}_{\text{flow}}(X)$} \label{linmodel}\\ 
    &~~~~~~~~~~~~\text{subject to}, ~0 \leq w_{\mathbf{F}_X \rightarrow X}, w_{A \rightarrow X} \leq 1, \forall A \in \mathbf{U}_X \label{lincons1} \\ &~~~~~~~~~~~~w_{\mathbf{F}_X \rightarrow X} + \underset{A \in \mathbf{U}_{X}}{\sum}w_{A \rightarrow X}=1 \label{lincons2}
    \end{align}
    $w_{\mathbf{F}_X \rightarrow X}$ and $w_{A \rightarrow X}$ are constrained between 0 and 1 since the objective of the mapper $f^\mathbf{w}$ is to capture the interaction between the fraction of the beliefs given by $w_{\mathbf{F}_X \rightarrow X}\mathbb{P}^{pa(X)_{\mathbf{F}_X}}_{\text{flow}}(X)$ and $\underset{A \in \mathbf{U}_{X}}{\bigcup}w_{A \rightarrow X}\mathbb{P}^{pa(X)_A}_{\text{flow}}(X)$ and approximate $P(x|pa(X))$. Eq. \ref{lincons1} and Eq. \ref{lincons2} ensure that the conditional probability axioms of $f^\mathbf{w}$ are satisfied.\\
    
    \item $f^\mathbf{w}=f^{\mathbf{w}_N}_N \circ ... \circ ~f^{\mathbf{w}_1}_1$ is composite function representing a N-layer neural network with $i^{th}$ layer having $M_i$ neurons and weights $\mathbf{w}_i$ capturing the non-linear combination of the inputs where,
    \begin{align}
    &\resizebox{0.72\textwidth}{!}{$f^\mathbf{w}(\mathbb{P}^ {pa(X)_{\mathbf{F}_X}}_{\text{flow}}(x), \underset{A \in \mathbf{U}_{X}}{\bigcup} \mathbb{P}^{pa(X)_A}_ {\text{flow}}(x))= f_N(...f_1(\mathbb{P}^{pa(X)_{\mathbf{F}_X}}_{\text{flow}}(x), \underset{A \in \mathbf{U}_{X}}{\bigcup} \mathbb{P}^{pa(X)_A}_{\text{flow}}(x)))$} \label{nonlinmodel} \\ \text{subject to,} ~&f_i: \mathbb{R}^{M_i} \rightarrow [0, 1]^{|X|}, \label{nonconst1}
    \\ &\resizebox{.4\textwidth}{!}{$\underset{x}{\sum} f^\mathbf{W}(\mathbb{P}^{pa(X)_{\mathbf{F}_X}}_{\text{flow}}(x), \underset{A \in \mathbf{U}_{X}}{\bigcup}\mathbb{P}^{pa(X)_A}_{\text{flow}}(x)) = 1$}  \label{nonconst2}
    \end{align}
    $f^\mathbf{W}$ captures the interaction between $\mathbb{P}^{pa(X)_{\mathbf{F}_X}}_{\text{flow}}(x)$ and $\underset{A \in \mathbf{U}_{X}}{\bigcup}\mathbb{P}^{pa(X)_A}_{\text{flow}}(x)$ and models $P(x|pa(X))$. Eq. \ref{nonconst1} and Eq. \ref{nonconst2} ensure that the conditional probability axioms of $f^\mathbf{W}$ are satisfied. One possibility is to use a softmax function for $f_N$ to ensure that the outputs of $f^\mathbf{W}$ satisfy probability axioms. 
\end{enumerate}

\section*{2. Proof of Theorem \ref{boundtheorem} \& Corollary \ref{boundtheoremcorol}}
\subsection*{Proof of Theorem \ref{boundtheorem}}
\begin{align}
&C_{\mathbf{S=s},Y=y} \\ \nonumber \\ &= \underset{\mathbf{s'}\in\mathbf{S} \backslash\mathbf{s}}{\sum} TE_{Y=y}(\mathbf{s,s'}) \mathbb{P}(\mathbf{s'}) \\ \nonumber \\ &= \underset{\mathbf{s'}\in \mathbf{S}\backslash\mathbf{s}}{\sum} [\mathbb{P}(y|do(\mathbf{s})) - \mathbb{P}(y|do(\mathbf{s'}))] \mathbb{P}(\mathbf{s'}) \\ \nonumber \\ &= \underset{\mathbf{s'}\in\mathbf{S}\backslash\mathbf{s}}{\sum} ~\mathbb{P}(\mathbf{s'})[\underset{\mathbf{v}_1\in\mathbf{V}\backslash Y}{\sum}\mathbb{P}(\mathbf{v}_1,y|do(\mathbf{s})) - \underset{\mathbf{v}_2\in\mathbf{V}\backslash Y}{\sum}\mathbb{P}(\mathbf{v}_2,y|do(\mathbf{s'}))] \nonumber \\ &[\mathbf{v}_1 ~\text{is consistent with} ~\mathbf{s} ~\text{and} ~\mathbf{v}_2 ~\text{is consistent with} ~\mathbf{s'}.] \\ \nonumber \\ &=\underset{\mathbf{s'}\in\mathbf{S} \backslash\mathbf{s}}{\sum}~\mathbb{P}(\mathbf{s'})[\underset{\mathbf{v}\in\mathbf{V}\backslash \{\textbf{S},Y\}}{\sum} ~\underset{V\in\mathbf{V}\backslash \{\mathbf{S},Y\},Y=y} {\prod} \mathbb{P}(v|pa(V))|_{\mathbf{s}}-\underset{\mathbf{v}\in\mathbf{V}\backslash \{\textbf{S},Y\}}{\sum} ~\underset{V\in\mathbf{V}\backslash \{\mathbf{S},Y\},Y=y} {\prod} \mathbb{P}(v|pa(V))|_{\mathbf{s'}}] \nonumber \\ &[\text{Definition} ~\ref{factorization}] \\ \nonumber \\
&=\underset{\mathbf{s'}\in\mathbf{S}\backslash\mathbf{s}}{\sum}~\mathbb{P}(\mathbf{s'})[\underset{\mathbf{v}\in\mathbf{V}\backslash \{\textbf{S},Y\}}{\sum} ~\underset{V\in\mathbf{V}\backslash \{\mathbf{S},Y\},Y=y} {\prod} f_{V}(\mathbb{P}^{pa(V)_ {\mathbf{F}_{V}}}_{\text{flow}}(v), \underset{A \in \mathbf{U}_{V}} {\bigcup}\mathbb{P}^{\mathbf{s}_A \lor pa(V)_A}
_{\text{flow}} ~(v))-\nonumber\\&\underset{\mathbf{v}\in\mathbf{V}\backslash \{\textbf{S},Y\}}{\sum} ~\underset{V\in\mathbf{V}\backslash \{\mathbf{S},Y\},Y=y} {\prod} f_{V} (\mathbb{P}^{pa(V)_ {\mathbf{F}_{V}}}_{\text{flow}}(v), \underset{A \in \mathbf{U}_{V}} {\bigcup}\mathbb{P}^{\mathbf{s'}_A \lor pa(V)_A}_{\text{flow}}(v))] \nonumber \\ &[\text{Theorem} ~\ref{cptdecom} ~\text{and} ~\text{Notation} ~\ref{notationor}] \\ \nonumber \\ \nonumber \\ &\leq \underset{\mathbf{s'}\in\mathbf{S}\backslash \mathbf{s}}{\sum}~\mathbb{P}(\mathbf{s'})[\underset{\mathbf{v}\in\mathbf{V}\backslash \{\textbf{S},Y\}}{\sum} ~\underset{V\in\mathbf{V}\backslash \{\mathbf{S},Y\},Y=y} {\prod} [f_{V}(\mathbb{P}^{pa(V)_{\mathbf{F}_{V}}}_{\text{flow}}(v),\underset{A \in \mathbf{U}_{V} \backslash B} {\bigcup}\mathbb{P}^{\mathbf{s}_A \lor pa(V)_A}_{\text{flow}}(v), \nonumber \\ &\mathbb{P}^{\mathbf{s}_B \lor pa(V)_B}_{\text{flow}} (v)=0) + \frac{\mathbb{P}^{\mathbf{s}_B \lor pa(V)_B}_ {\text{flow}}(v)}{\mathbb{P}(v,pa(V))|_{\mathbf{s}}}\mu_{B \rightarrow V}] -\underset{\mathbf{v}\in\mathbf{V}\backslash \{\textbf{S},Y\}}{\sum} ~\underset{V\in\mathbf{V}\backslash \{\mathbf{S},Y\},Y=y} {\prod} \nonumber\\&[f_{V}(\mathbb{P}^{pa(V)_{\mathbf{F}_{V}}}_{\text{flow}}(v),\underset{A \in \mathbf{U}_{V} \backslash B} {\bigcup}\mathbb{P}^{\mathbf{s'}_A \lor pa(V)_A}_ {\text{flow}}(v),\mathbb{P}^{\mathbf{s'}_B \lor pa(V)_B}_{\text{flow}}(v)=0)-\frac{\mathbb{P}^{\mathbf{s'}_B \lor pa(V)_B}_{\text{flow}}(v)}{\mathbb{P}(v,pa(V))|_{\mathbf{s'}}}\mu_{B \rightarrow V}]] \nonumber \\ &[\text{Definition \ref{edgeunfair}, property that} ~|.| \geq 0, ~\text{and} ~\text{Notation} ~\ref{notationor}]
\end{align}

\begin{align}
&\leq \underset{\mathbf{s'}\in\mathbf{S} \backslash\mathbf{s}}{\sum}~\mathbb{P}(\mathbf{s'})[\underset{\mathbf{v}\in\mathbf{V}\backslash \{\textbf{S},Y\}}{\sum} ~\underset{V\in\mathbf{V}\backslash \{\mathbf{S},Y\},Y=y} {\prod} [f_{V}(\mathbb{P}^{pa(V)_{\mathbf{F}_{V}}}_{\text{flow}}(v),\underset{A \in \mathbf{U}_{V}} {\bigcup}\mathbb{P}^{\mathbf{s}_A \lor pa(V)_A}_{\text{flow}}(v)=0) \nonumber \\ &+ \underset{A \in \mathbf{U}_{V}}{\sum}\frac{\mathbb{P}^{\mathbf{s}_A \lor pa(V)_A}_{\text{flow}}(v)} {\mathbb{P}(v,pa(V))|_{\mathbf{s}}} \mu_{A \rightarrow V}] -\underset{\mathbf{v}_2\in\mathbf{V} \backslash \{\textbf{S},Y\}}{\sum} ~\underset{V\in\mathbf{V} \backslash \{\mathbf{S},Y\},Y=y}{\prod}[f_{V}(\mathbb{P}^{pa(V)_{\mathbf{F}_{V}}}_{\text{flow}}(v), \nonumber\\&\underset{A \in \mathbf{U}_{V}} {\bigcup}\mathbb{P}^{\mathbf{s'}_A \lor pa(V)_A}_{\text{flow}} (v)=0)- \underset{A\in \mathbf{U}_{\mathbf{V}}}{\sum} \frac{\mathbb{P}^{\mathbf{s'}_A \lor pa(V)_A}_{\text{flow}} (v)}{\mathbb{P}(v, pa(V))|_{\mathbf{s'}}} \mu_{A \rightarrow V}]] \nonumber \\ &[\text{Recursively apply previous step for every A} \in \mathbf{U}_{V} ~\text{and} ~\text{Notation} ~\ref{notationor}] \\ \nonumber \\ \nonumber \\ \nonumber \\ &\leq \underset{\mathbf{s'}\in\mathbf{S}\backslash\mathbf{s}}{\sum}~\mathbb{P}(\mathbf{s'})[\underset{\mathbf{v}\in\mathbf{V}\backslash \{\textbf{S},Y\}}{\sum} ~\underset{V\in\mathbf{V}\backslash \{\mathbf{S},Y\},Y=y} {\prod} \underset{~A \in \mathbf{U}_{V}} {\sum}\frac{\mathbb{P}^{\mathbf{s}_A \lor pa(V)_A}_{\text{flow}} (v)}{\mathbb{P}(v,pa(V))|_{\mathbf{s}}}\mu_{A \rightarrow V} +\nonumber\\&\underset{\mathbf{v}\in\mathbf{V}\backslash \{\textbf{S},Y\}}{\sum} ~\underset{V\in\mathbf{V}\backslash \{\mathbf{S},Y\},Y=y} {\prod}\underset{~A\in \mathbf{U}_{\mathbf{V}}} {\sum}\frac{\mathbb{P}^{\mathbf{s'}_A \lor pa(V)_A}_{\text{flow}}(v)}{\mathbb{P}(v, pa(V))|_{\mathbf{s'}}} \mu_{A \rightarrow V}] \\ \nonumber \\ \nonumber \\ &\leq \underset{\mathbf{s'}\in\mathbf{S}\backslash\mathbf{s}}{\sum}~\mathbb{P}(\mathbf{s'})[\underset{v\in\mathbf{V}\backslash \{\textbf{S},Y\}}{\sum} ~\underset{V\in\mathbf{V}\backslash \{\mathbf{S},Y\},Y=y} {\prod} \underset{~A \in \mathbf{U}_{V}} {\sum}\left[\frac{\mathbb{P}^{\mathbf{s}_A \lor pa(V)_A}_{\text{flow}} (v)}{\mathbb{P}(v,pa(V))|_{\mathbf{s}}}+\frac{\mathbb{P}^{\mathbf{s'}_A \lor pa(V)_A}_{\text{flow}}(v)}{\mathbb{P}(v, pa(V))|_ {\mathbf{s'}}}\right] \mu_{A \rightarrow V}] \label{cumuppb1} 
\end{align}

Thus, 
\begin{align}
C_{\mathbf{S=s},Y=y} &\leq C^{\text{upper}}_{\mathbf{S=s},Y=y}\hfill ~~~~[\text{From Eq.} ~\ref{cumuppb1} ~\text{and Eq.} ~\ref{cumuppb2}] \\ C_{\mathbf{S=s},Y=y} &\geq -C^{\text{upper}}_{\mathbf{S=s},Y=y} \hfill ~~~~[\text{Similar proof}] \\ \therefore |C_{\mathbf{S=s},Y=y}| &\leq C^{\text{upper}}_{\mathbf{S=s},Y=y} \label{thm4statement} ~~~~~~~~~~~~~\blacksquare  \\ \nonumber 
\end{align} 

\subsection*{Proof of Corollary \ref{boundtheoremcorol}}
When edge unfairness $\mu_{e,\mathbb{G}}=0, ~\forall e$ from $\mathbf{S}$,
\begin{align}
&C^{\text{upper}}_{\mathbf{S=s},Y=y} = 0 \nonumber \\ &[\text{Edge unfairness} ~\mu_{e,\mathbb{G}}=0 ~\forall e ~\text{from} ~\mathbf{S} ~\text{and Eq.} ~\ref{cumuppb2}] \label{cumuppzero} \\ \nonumber \\ &C_{\mathbf{S=s},Y=y}=0 \nonumber \\ &[\text{Eq.} ~\ref{thm4statement} ~\text{and Eq.} ~\ref{cumuppzero}]  ~~~~~~~~~~\blacksquare
\end{align}

\section*{3. Experiments - Additional Details} \label{app_table}
The values taken by each of the node in the causal graph \ref{main_graph} are shown in Table \ref{att_table}.
\begin{table}[h!]
    \centering
    \caption{Nodes and their Values.}
    \label{att_table}
    \resizebox{0.8\linewidth}{!}{\begin{tabular}{c|c}
    \textbf{Node} & \textbf{Values}\\
        \hline
        Race $R$ & African American($0$), Hispanic($1$) and White($2$) \\
        Gender $G$ & Male($0$), Female($1$) and Others($2$) \\
        Age $A$ & Old ($0$)($>$35y) and Young ($1$) ($\leq$ 35y) \\
        Literacy $L$ & Literate ($0$) and Illiterate ($1$) \\
        Employment $E$ & Not Employed ($0$) and Employed ($1$) \\
        Bail Decision $J$ & Bail granted ($0$) and Bail rejected ($1$) \\
        Case History $C$ & Strong ($0$) and Weak criminal history ($1$)
    \end{tabular}}
\end{table}

\subsection*{3.1 Edge Unfairness is an Edge Property}
\label{unfairedgeproperty}
We investigate that the edge unfairness depends on the parameters of the edge and not on the specific values of the attributes. \\
\begin{figure}[h!]
\centering
\subfigure[]{\includegraphics[width=0.75\linewidth, height=4.75cm]{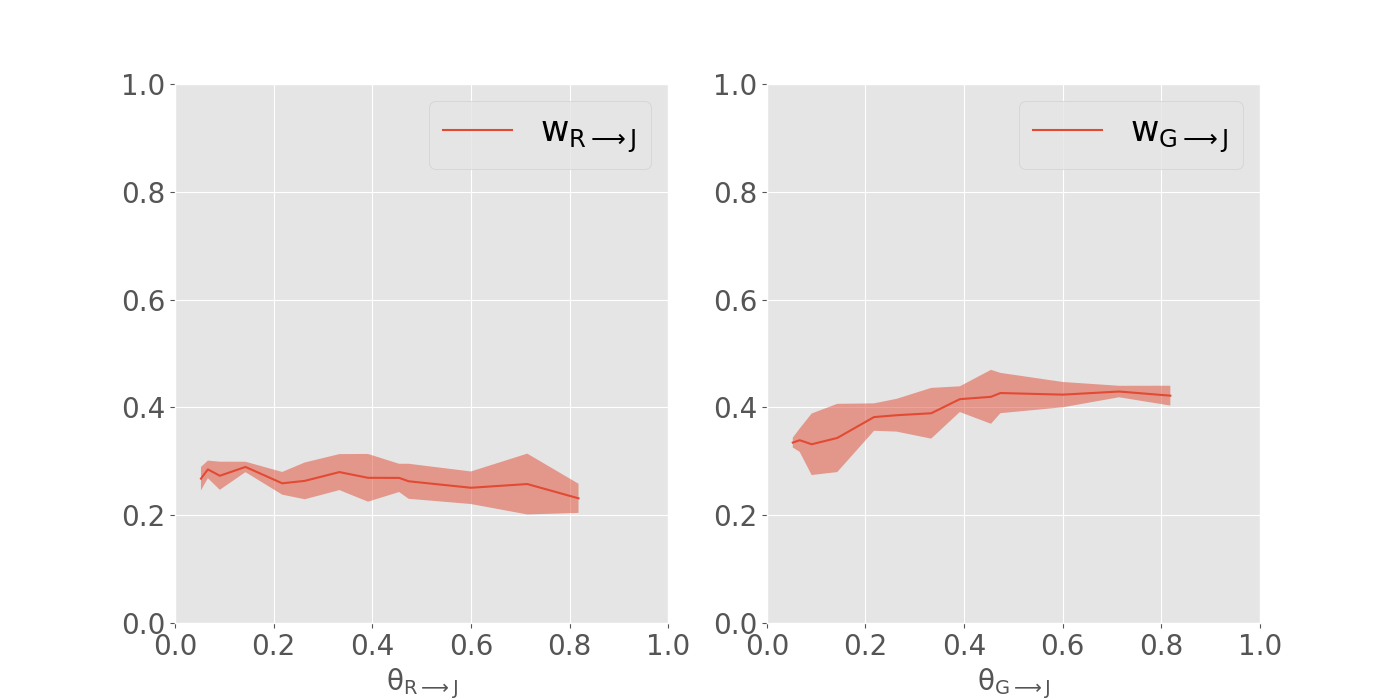}} \\
\subfigure[]{\includegraphics[width=0.75\linewidth,height=4.75cm]{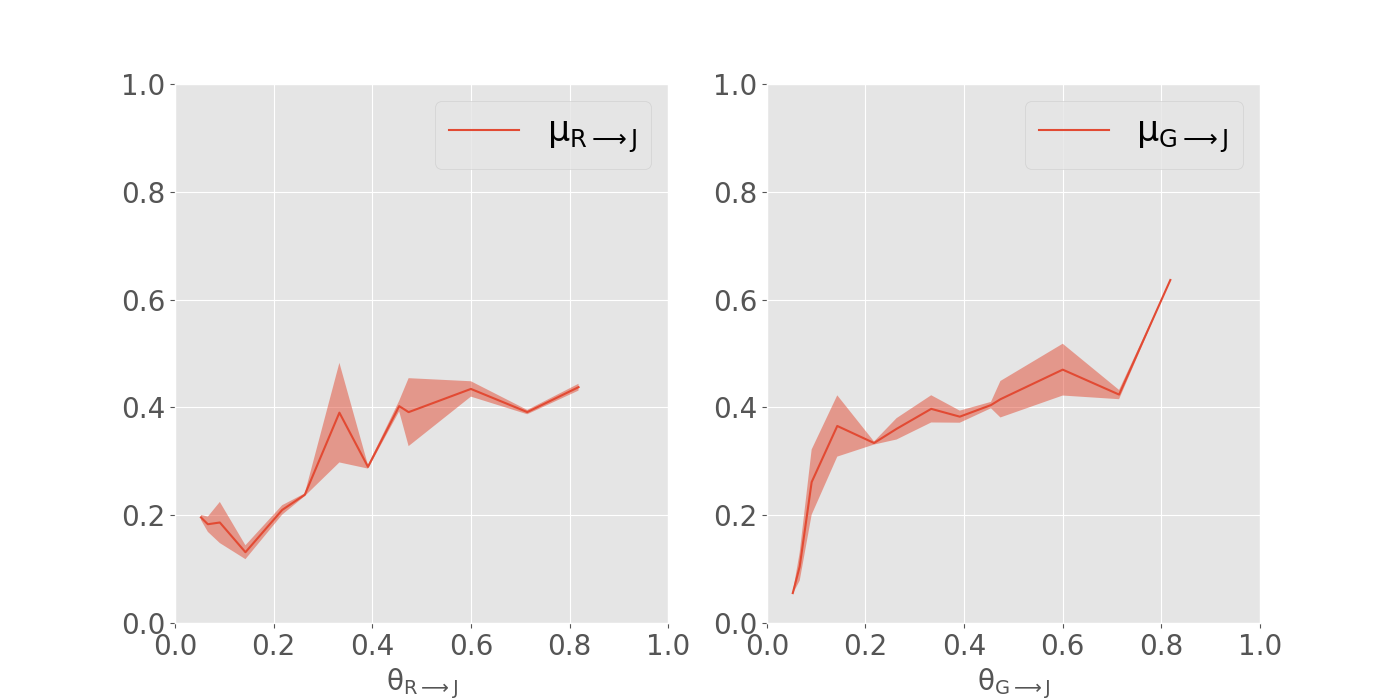}}
\caption{Edge unfairness is a property of the edge because there is minimal variation in edge unfairness for a specific $\theta_{e}$. (a) $w^*_{R\rightarrow J}$ vs. $\theta_{R\rightarrow J}$ and $w^*_{G\rightarrow J}$ vs. $\theta_{G\rightarrow J}$ for linear model. (b) $\mu_{R\rightarrow J}$ vs. $\theta_{R\rightarrow J}$ and $\mu_{G\rightarrow J}$ vs. $\theta_{G\rightarrow J}$ for non-linear model.\\} \label{theta_weight}
\end{figure}

\textbf{\textit{Inference:}} When a linear model is used, $\mathbf{w}^*$ is observed to be insensitive to the specific values taken by the nodes as there is minimal variation in $\mathbf{w}^*_{e}$ for any fixed $\theta_{e}$ as shown in Fig. \ref{theta_weight}(a). $w^*_{R\rightarrow J}$ was observed to be in the range $[0.2, 0.3]$ for different $\theta_{R\rightarrow J}$. A small deviation in $w^*_{R\rightarrow J}$ shows that $w^*_{R\rightarrow J}$ depends only on $\theta_{R\rightarrow J}$ and not on the specific values taken by the nodes. Since edge unfairness in an edge, say $R \rightarrow J$, is $\mu_{R \rightarrow J}=|Pa(J)|w_{R \rightarrow J}$ in the linear model setting, it indicates that edge unfairness is also insensitive to the specific values taken by nodes and hence is a property of the edge. Similarly for the non-linear model, edge unfairness $\mu_{e}$ is insensitive to the specific values taken by the nodes as there is minimal variation in $\mu_{e}$ for any fixed $\theta_{e}$ as observed from Fig. \ref{theta_weight}(b). For instance, $\mu_{R\rightarrow J}$ obtained in the models with $\theta_{R\rightarrow J} = 0.5$ are in the range $[0.35, 0.43]$. A similar observation can be made for $w^*_{G\rightarrow J}$ and $\mu_{G\rightarrow J}$ in Fig. \ref{theta_weight}(a) and \ref{theta_weight}(b) respectively. We also analyze the \textit{MSE} for both the linear and non-linear settings in Supplementary material.

\subsection*{3.2 Linear and Non-linear model comparison} \label{app_lin_nonlin}
To validate the benefits of a non-linear model, the \textit{MSE}s between the \textit{CPT}s for bail decision $\mathbb{P}(J|Pa(J))$ and its functional approximation $f^\mathbf{w}$ were recorded for these settings:
\begin{enumerate}
    \item \textit{MSE}s $e^L_J$ calculated when $f^\mathbf{w}$ is approximated using a linear model (Eq. \ref{linmodel})
    \item \textit{MSE}s $e^{NL}_J$ calculated when $f^\mathbf{w}$ is approximated using a non-linear model (Eq. \ref{nonlinmodel})
\end{enumerate}

\begin{figure}[h!]
\centering
\includegraphics[width=0.55\linewidth]{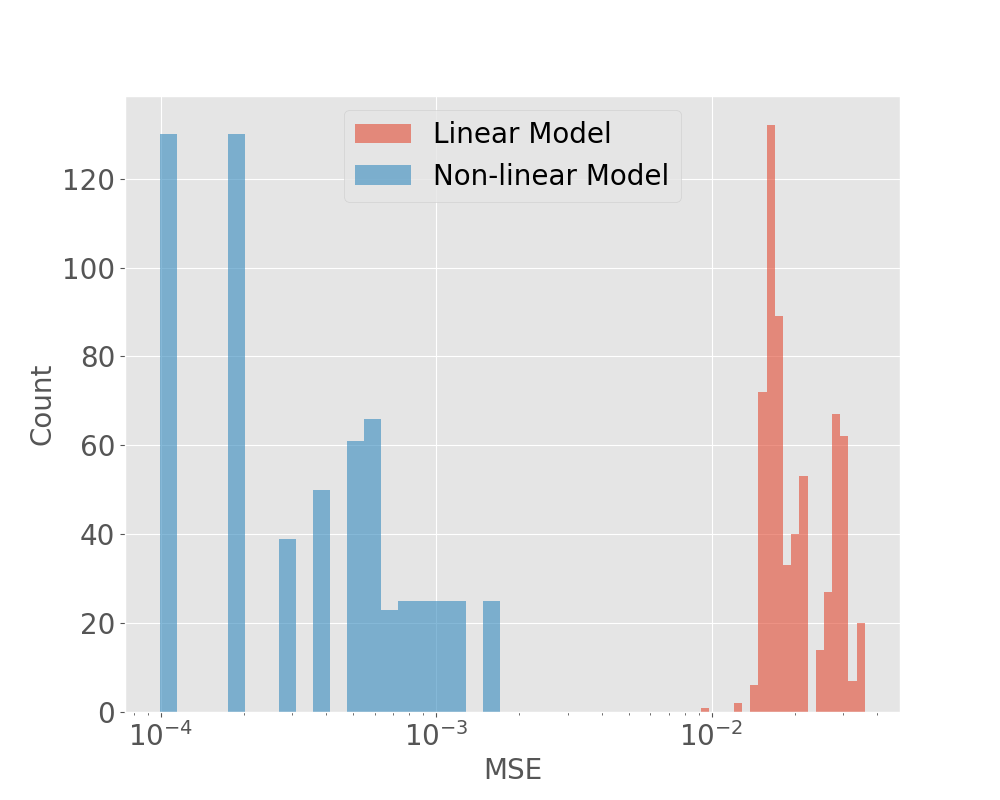}
\caption{Histogram for \textit{MSE} by using a linear model shown in red and using a non-linear model shown in blue for $625$ different models (discussed in Section \ref{causal_exp}).\\} \label{lin_nonlin}
\end{figure}

\textbf{Inference:} Distributions of $e^L_J$ and $e^{NL}_J$ are plotted in Fig. \ref{lin_nonlin}. Here, the maximum value of $e^L_J$ shown in the red bar is obtained above $0.01$ and its values mostly lie in the range $(0.01,0.02)$. On the other hand, $e^{NL}_J$ shown in blue bars is distributed in the range $(0.0001,0.001)$ with the maximum value of $e^{NL}_J$ obtained around $0.002$. Hence, a non-linear model like a neural network to approximate $f^\mathbf{w}$ is a better choice because the \textit{MSE}s distribution lies in the lower error range.
\end{document}